\pdfoutput=1

\documentclass[11pt]{article}

\usepackage{EMNLP2022}

\usepackage{times}
\usepackage{latexsym}
\usepackage{graphicx} 
\graphicspath{ {images/} }
\usepackage{xcolor}
\usepackage{listings}
\usepackage[T1]{fontenc}

\usepackage[utf8]{inputenc}

\usepackage{microtype}

\usepackage{inconsolata}

\usepackage[super]{nth}

\colorlet{punct}{red!60!black}
\definecolor{background}{HTML}{EEEEEE}
\definecolor{delim}{RGB}{20,105,176}
\colorlet{numb}{magenta!60!black}
\lstdefinelanguage{json}{
    basicstyle=\normalfont\ttfamily,
    numbers=left,
    numberstyle=\scriptsize,
    stepnumber=1,
    numbersep=8pt,
    showstringspaces=false,
    breaklines=true,
    frame=lines,
    backgroundcolor=\color{background},
    literate=
     *{0}{{{\color{numb}0}}}{1}
      {1}{{{\color{numb}1}}}{1}
      {2}{{{\color{numb}2}}}{1}
      {3}{{{\color{numb}3}}}{1}
      {4}{{{\color{numb}4}}}{1}
      {5}{{{\color{numb}5}}}{1}
      {6}{{{\color{numb}6}}}{1}
      {7}{{{\color{numb}7}}}{1}
      {8}{{{\color{numb}8}}}{1}
      {9}{{{\color{numb}9}}}{1}
      {:}{{{\color{punct}{:}}}}{1}
      {,}{{{\color{punct}{,}}}}{1}
      {\{}{{{\color{delim}{\{}}}}{1}
      {\}}{{{\color{delim}{\}}}}}{1}
      {[}{{{\color{delim}{[}}}}{1}
      {]}{{{\color{delim}{]}}}}{1},
}

\title{Masader Plus: A New Interface for Exploring +500 Arabic NLP Datasets}

\author{
Yousef Altaher $^1$~~~ 
Ali Fadel  $^2,$\thanks{\phantom{...}This work is not related to Amazon}~~~ 
Mazen Alotaibi $^3$~~~ 
Mazen Alyazidi $^4$~~~ 
Mishari Al-Mutairi $^5$~~~
\\
\textbf{
Mutlaq Aldhbuiub $^6$~~~
Abdulrahman Mosaibah $^7$~~~ 
Abdelrahman Rezk $^8$~~~  
Abdulrazzaq Alhendi $^9$~~~}
\\
\textbf{
Mazen Abo Shal $^4$~~~ 
Emad A. Alghamdi $^{10}$~~~ 
Maged S. Alshaibani $^{11}$~~~
Jezia Zakraoui $^{12}$~~~}
\\
\textbf{
Wafaa Mohammed $^{13}$~~~ 
Kamel Gaanoun $^{14}$~~~ 
Khalid N. Elmadani $^{15}$~~~ 
Mustafa Ghaleb $^{16}$~~~}
\\
\textbf{
Nouamane Tazi $^{17}$~~~ 
Raed Alharbi  $^{18}$~~~
Maraim Masoud  $^{3}$~~~
Zaid Alyafeai$^{11,\gamma}$~~~}
\smallskip 
\\
$^1$ King's College London, United Kingdom
$^2$Amazon, Jordan
$^3$Independent Researcher
\\
$^4$Independent Software Developer, Saudi Arabia
$^5$Independent Software Developer and Designer, Saudi Arabia
\\
$^6$Independent Software Engineer, Saudi Arabia
$^7$University of Bahrain, Bahrain
$^8$ IIT Madras, India,
\\
$^9$Dasman Diabetes Institute, Kuwait
$^{10}$King Abdulaziz University, AILLA Lab, Saudi Arabia
\\
$^{11}$KFUPM, Saudi Arabia
$^{12}$Independent Researcher, Qatar
$^{13}$University of Tübingen, Germany
\\
$^{14}$INSEA, Morocco
$^{15}$University of Cape Town, South Africa
$^{16}$KFUPM, IRC-ISS, Saudi Arabia
\\
$^{17}$Hugging Face, Inc
$^{18}$Saudi Electronic University, Saudi Arabia
\\
$^{\gamma}$ \nolinkurl{g201080740@kfupm.edu.sa}
}
\begin{document}
\maketitle

\begin{abstract}
Masader \citep{alyafeai2021masader} created a metadata structure to be used for cataloguing Arabic NLP datasets. However, developing an easy way to explore such a catalogue is a challenging task. In order to give the optimal experience for users and researchers exploring the catalogue, several design and user experience challenges must be resolved. Furthermore, user interactions with the website may provide an easy approach to improve the catalogue. In this paper, we introduce \emph{Masader Plus}, a web interface for users to browse Masader. We demonstrate data exploration, filtration, and a simple API that allows users to examine datasets from the backend. Masader Plus can be explored using this link \url{https://arbml.github.io/masader}. A video recording explaining the interface can be found here \url{https://www.youtube.com/watch?v=SEtdlSeqchk}.
\end{abstract}

\section{Introduction}
Recently, much research work targeted different aspects related to the processing of Arabic and its dialects such as morphological analysis, resource building, machine translation, etc. However, according to \cite{GUELLIL2021497}, most research  concentrated on building resources (lexicon, corpora, datasets). Arguably, the growth in NLP research also brings growth in datasets, which presents substantial challenges for potential users in terms of resource retrieval, access, and re-use. However, research efforts that addressed metadata sourcing for Arabic, are available either as a review \cite{Zaghouani2017CriticalSO}, or as a public catalogue \cite{alyafeai2021masader} only. An intuitive user interface (UI) design \cite{intuitiveUI-Bernal} capturing all required functionalities such as dynamic search and filtering, sorting functions, descriptive statistics, and data visualization need to be implemented to target different audience like researchers, social scientists, and regular users. 
The primary goal of this work is to enhance the work of \cite{alyafeai2021masader} on both contextual and visual features of datasets. Masader Plus ensures up-to-date availability of dataset’s metadata. Furthermore, the interface provides researchers with a set of user controls for filtering, refining, and visualizing depending on metadata qualities to aid in user exploration of the metadata. Masader Plus is completely open source and available with \texttt{GPL-3.0} license at \url{https://github.com/arbml/masader}. 
We summarize our contributions as the following:

\begin{enumerate}
    \item API endpoints that support search, filtration, indexing, and reporting discussed in \ref{sec:apis}.
    \item Search page with advanced filtration detailed in section \ref{sec:filter}
    \item Metadata visualization by cluster, task, domain etc. detailed in section \ref{sec:vis}
\end{enumerate}

In the following section, we highlight the related works. Then, in section \ref{sec:system}, we outline our system description, which is followed by a presentation of the system architecture and a demonstration of the system features in section \ref{sec:arc}. We, then, present the community contribution effort in section \ref{sec:com}. Section  \ref{sec:eth} discusses the ethics and a broad impact statement. Finally, we conclude the paper with a conclusion and future work in section \ref{sec:conc}.



\begin{figure*}[htp!]
\centering
\includegraphics[width=\textwidth]{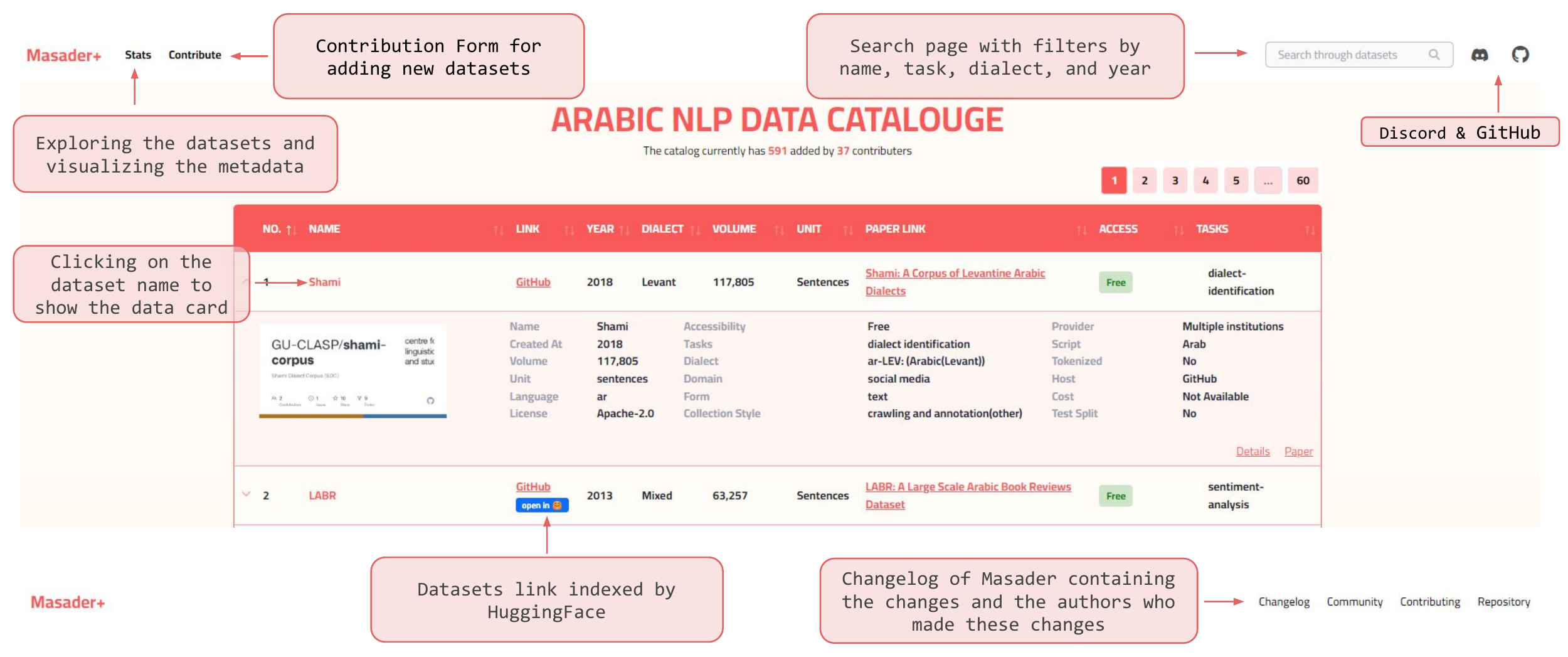}
\caption{Frontend components.}
\label{fig:frontend}
\end{figure*}

\section{Related Work}
\label{sec:lit}
There have been continuous efforts to collect, categorize, and group NLP datasets in different languages with the goal of facilitating research practice. 
Masader catalogue follows a community driven approach to adding and documenting datasets, which is similar to the approach of \cite{lhoest2021datasets} in the creation of \textit{Datasets} library.
HuggingFace \textit{Datasets} \footnote{\url{https://huggingface.co/datasets}} is a community-driven, open-source library that standardizes the processing, distribution, and documentation of NLP datasets. It facilitates different stages of working with a dataset; which includes loading the whole dataset, accessing its features schema and metadata, slicing, batching, and parallel processing. However the number of Arabic NLP datasets is below 100 as accessed on August 1st, 2022.\\
While \textit{Datasets} does not host most of the underlying raw datasets, there are other frameworks that offer cloud-based storage of large databases of datasets. These efforts have mainly emerged with the recent development of deep learning technologies. Examples include \cite{tfdatasets} and \cite{torch} which create cloud based repositories of datasets. They store the datasets in a common cloud format. \\
Lanfrica \footnote{\url{https://lanfrica.com/records}} is another community specific search engine that aims to create a centralised hub for all African language resources to make it easy for NLP researchers, linguists, government officials, and general users to access this data \cite{emezue2020lanfrica}. They provide a small number of annotations for the metadata. \\
Other language-specific catalogues include \cite{german} for German, \cite{french} for French, \cite{indic} for Indic NLP resources, and \cite{tamil} a Tamil NLP catalogue. These catalogues only offer a list of resources without indicating their metadata, they also do not offer any filtration capabilities which makes it difficult to navigate through and explore the resources.

\section{System Description}
\label{sec:system}
Masader Plus system consists of four main workflows that are related to the users, contributors, administrators, and data. The four workflows are illustrated in Figure \ref{fig:system_design}. The list below explains these workflows in their natural order and shows how they are interacting together to deliver up-to-date datasets metadata:

\begin{figure*}[htp!]
\centering
\includegraphics[width=0.8\textwidth]{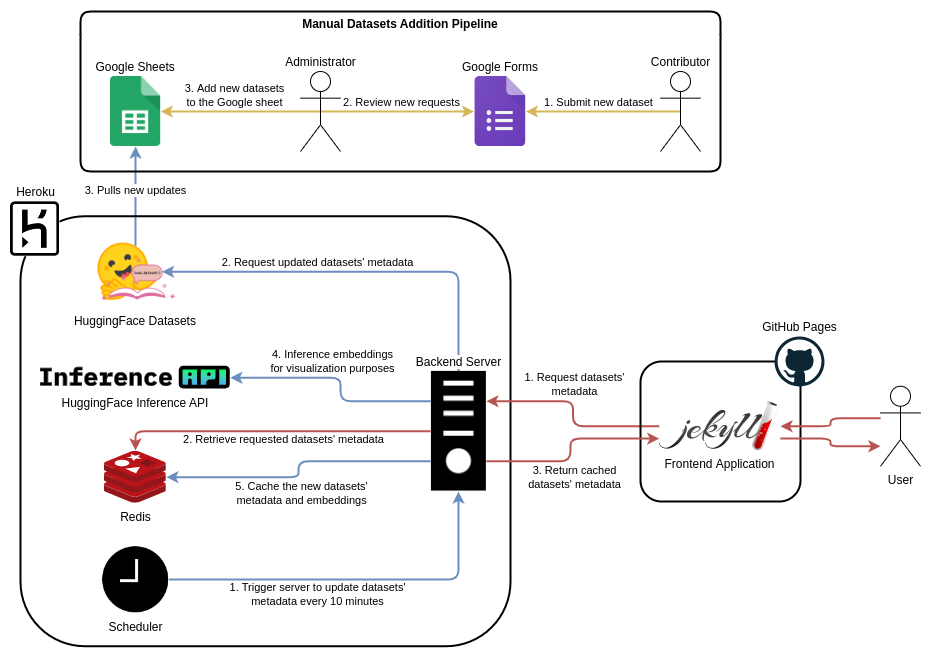}
\caption{System design and workflow illustration for the users, contributors, administrators, and data.}
\label{fig:system_design}
\end{figure*}

\begin{itemize}
    \item \textbf{Contributors Workflow}: To contribute a new dataset to Masader, contributors need to fill \href{https://forms.gle/JnMrJjHumT6ktK9cA}{Masader Google Form} and submit it with the required information.
    \item \textbf{Administrators Workflow}: After receiving the contribution submission, administrators will review the submission and either update it if there are any issues, or approve it and update the Google Sheet with the new dataset metadata.
    \item \textbf{Data Workflow}: In the backend side, the server will be triggered every 10 minutes automatically to get the newly updated datasets metadata through the \href{https://huggingface.co/datasets/arbml/masader}{Masader dataset}. After getting the updated datasets metadata, the server will request HuggingFace Inference APIs to compute the new datasets embeddings using \href{https://huggingface.co/sentence-transformers/all-MiniLM-L6-v2}{\texttt{all-MiniLM-L6-v2}} model from Sentence Transformers. Using the newly computed embeddings, the server will compute the new clustring information for all datasets and cache the full datasets metadata in Redis to respond to requests in a constant time.
    \item \textbf{Users Workflow}: Masader Plus end users can use the frontend application to explore the available up-to-date datasets metadata through: 1) Tabular view that can be filtered to find the required dataset; 2) The available datasets metadata statistics to understand the distribution of the datasets and get more useful insights. 
\end{itemize}

\section{Architecture and Features}
\label{sec:arc}
Masader Plus is a web application that consists of a backend layer and a frontend layer. 


\paragraph{Backend.} Masader Plus backend application is hosted on Heroku hosting service\footnote{\url{https://www.heroku.com/}} and it is built on top of Flask framework\footnote{\url{https://flask.palletsprojects.com/en/2.1.x/changes/\#version-2-1-2}} that supports rapid development. For storage, Masader Plus is relying on Redis\footnote{\url{https://github.com/redis/redis-py/releases/tag/v4.3.4}} as a caching service for the computed datasets metadata. To do the heavy computations for embeddings extraction, Masader Plus is utilizing HuggingFace Inference API\footnote{\url{https://huggingface.co/inference-api}} to calculate the embeddings for each dataset in the data store. Lightweight computations like clustering are computed on the same backend server.
\paragraph{Frontend.} The application frontend is implemented using HTML, JavaScript, CSS and Bootstrap v5. We also use Ruby, Bundler and Jekyll as a templating language. To run the server locally, contributors can use the contributing page \footnote{\url{https://github.com/arbml/masader/blob/main/CONTRIBUTING.md}}. Figure \ref{fig:frontend} depicts the main interface of Masader Plus after running the server locally.

\subsection{Masader APIs}
\label{sec:apis}

Masader APIs \footnote{\url{https://github.com/arbml/masader-webservice}} provide a set of endpoints to simplify the access to the metadata:

\begin{itemize}
    \item \href{http://apis.masader.arbml.org/datasets/schema}{\texttt{\textbf{/datasets/schema}}}: Returns a list of available features for the datasets. Example response:
    \begin{lstlisting}[language=json,basicstyle=\small]
["Name", "Year", "Unit", ...]
    \end{lstlisting}
    \item \href{http://apis.masader.arbml.org/datasets}{\texttt{\textbf{/datasets}}}: Returns a list of available datasets based on the passed \texttt{query} and the requested \texttt{features} parameters.
    Using the \texttt{query} parameter, a \texttt{filtration} query will be applied on the datasets before selecting the required features and returning the output (e.g. \href{http://apis.masader.arbml.org/datasets?query=Year>2003 and Year<2008 and Unit=='tokens'}{\texttt{query=Year>2003 and Year<2008 and Unit=={\textquotesingle}tokens\textquotesingle}}). We use the Pandas query language, for more information see \href{https://pandas.pydata.org/docs/reference/api/pandas.DataFrame.query.html}{here}.
    Using \texttt{features} parameter, the list of features for each dataset will be filtered based on the passed value (e.g. \href{http://apis.masader.arbml.org/datasets?features=Name,Year,Unit}{\texttt{features=Name,Year,Unit}}). Example response:
    \begin{lstlisting}[language=json,basicstyle=\small]
[
    {
        "Name": "Shami",
        "Unit": "sentences",
        "Year": 2018
    },
    ...
]
    \end{lstlisting}
    \item \href{http://apis.masader.arbml.org/datasets/2}{\texttt{\textbf{/datasets/[index]}}}: Returns a specific dataset from the available datasets based on its \texttt{index}. Example response:
    \begin{lstlisting}[language=json,basicstyle=\small]
{
    "Name": "LABR",
    "Year": 2018,
    "Dialect":"mixed",
    ...
}
    \end{lstlisting}
    \item \href{http://apis.masader.arbml.org/datasets/tags}{\texttt{\textbf{/datasets/tags}}}: Returns the unique values of the requested \texttt{features}. Using \texttt{features} parameter, the list of features will be filtered based on the passed value (e.g. \href{http://apis.masader.arbml.org/datasets/tags?features=Dialect,Year}{\texttt{features=Dialect,Year}}). Example response:
    \begin{lstlisting}[language=json,basicstyle=\small]
{
    "Dialect": [
        "Algeria",
        "Bahrain",
        ...
    ],
    "Year": [
        2001,
        2002,
        ...
    ]
    ...
}
    \end{lstlisting}
\end{itemize}

\subsection{Data Cards}

\begin{figure}[h]
\centering
\includegraphics[width=0.40\textwidth]{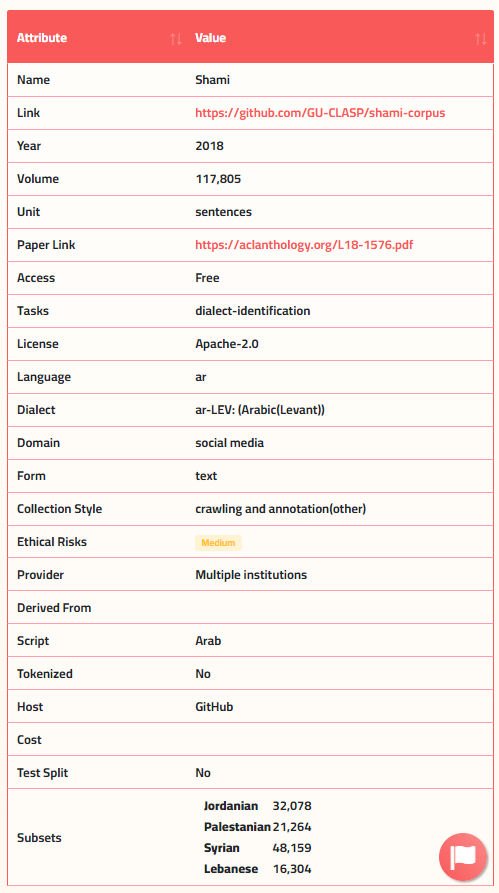}
\caption{An example data card.}
\label{fig:datacard1}
\end{figure}

Data Cards are defined as \emph{structured summaries of essential facts about various aspects of ML datasets needed by stakeholders across a dataset's life-cycle for responsible AI development} \cite{datacardsPushkarna}. In our project each data card contains documentation about its host, year of publication, access type, etc. refer to the Masader paper \cite{alyafeai2021masader} for more information about the metadata. Besides, a reporting functionality is implemented to report any issue on the data cards either anonymously or using a GitHub account. Figure \ref{fig:datacard1} presents an example of the data card page on the Masader Plus website. 

\subsection{Feature \#1: Advanced Filtration}
\label{sec:filter}
\begin{figure}[h]
\centering
\includegraphics[width=0.40\textwidth]{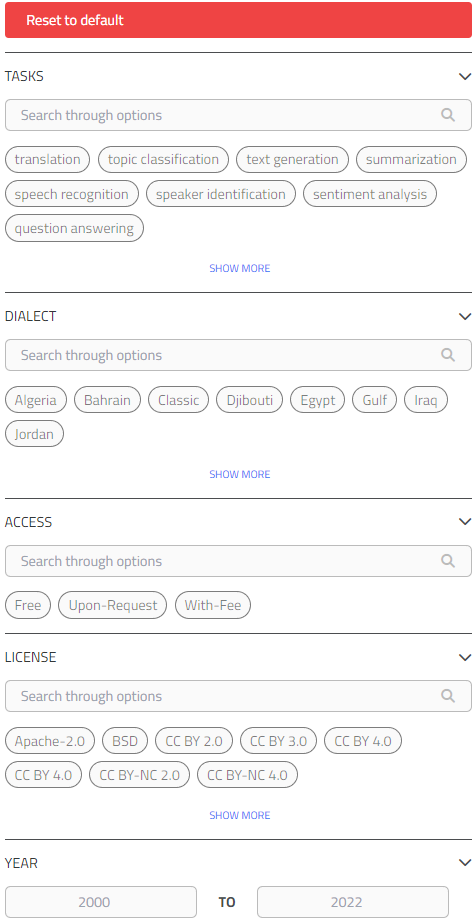}
\caption{Filtration Sidebar.}
\label{fig:filter}
\end{figure}

We want to allow users to easily explore the datasets by searching with advanced conditions. Given there are many conditions that could be applied, we added some endpoints to the web service to make this possible (see subsection \ref{sec:apis}). To allow a simple usage of such \emph{endpoints}, we created a simple search page that could be used to extract the datasets with specific conditions. In Figure \ref{fig:filter}, we highlight the main filtration conditions in the interface:
\begin{itemize}
    \item \textbf{Tasks}: filter by NLP tasks. We show the first 20 most used tasks at the beginning.
    \item \textbf{Dialect} allow filtering by either region or a given country dialect. 
    \item \textbf{Access} filtration by how easy the access of the dataset is. 
    \item \textbf{License} Licenses for accessing the datasets. 
    \item \textbf{Year} allow filtering by a given year range. The current Masader catalogue contains datasets from 2000 to 2022. 
\end{itemize}

\subsection{Feature \#2: Datasets and Metadata Visualization}
\label{sec:vis}
\begin{figure}[h]
\centering
\includegraphics[width=0.5\textwidth]{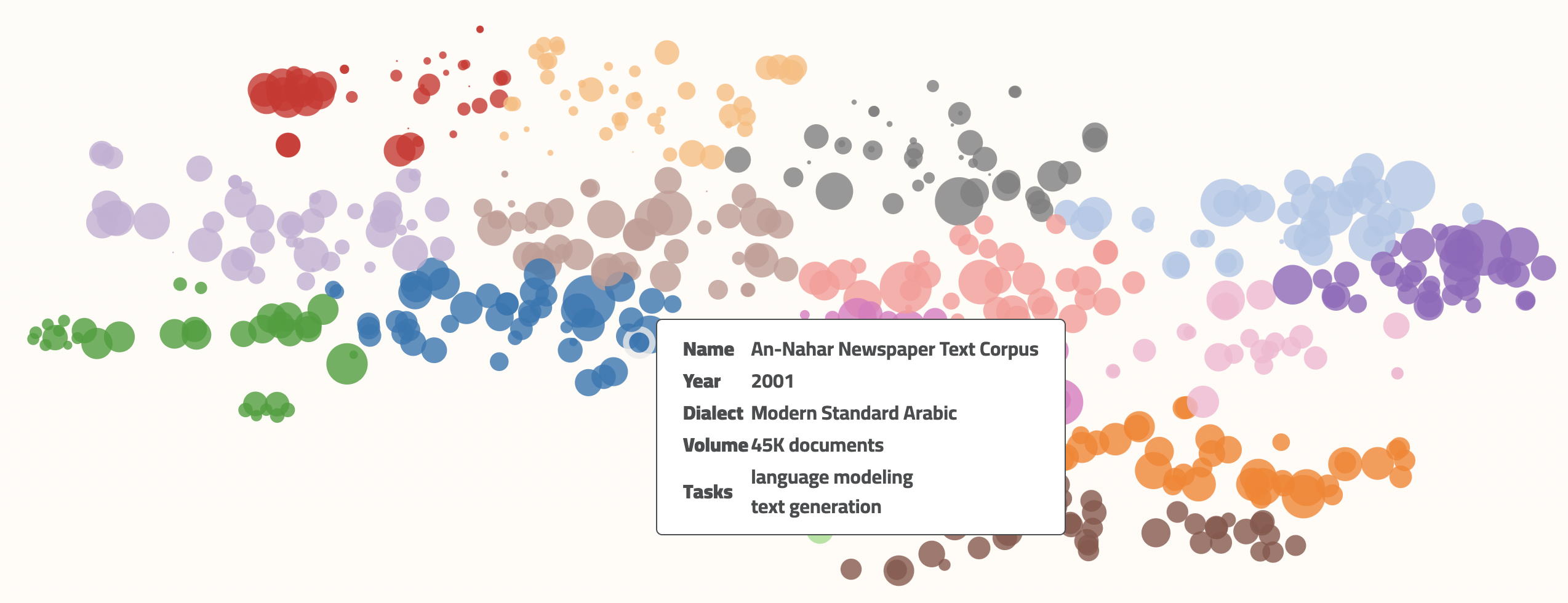}
\caption{The clusters graph when hovering over a dataset node.}
\label{fig:clusters}
\end{figure}

The Stats page contains various visuals outlining the specific features of the dataset in different formats. The first graph (as shown in Figure \ref{fig:clusters}) presents the datasets as nodes in the embedding space clustered using K-Means. The embeddings were encoded using the Name, Description and Abstract metadata of the datasets. Other graphs, show the metadata like the Host, Year, Access, Tasks, Domain, Licenses, Dialects, Forms, Venues, Ethical Risks, and Scripts. Such metadata apart from the dialects, are visualized using bar charts, while a doughnut-chart is used for the dialects. 

\section{Community Contributions}
\label{sec:com}
\begin{figure*}[h]
\begin{center}
    \includegraphics[width=0.9\textwidth]{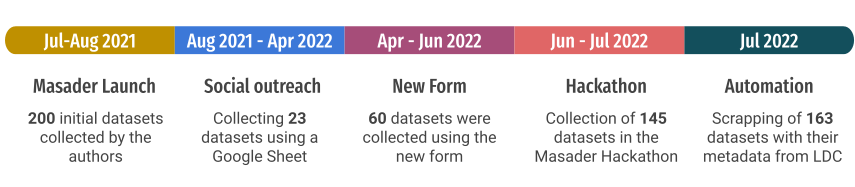} 
  \caption{Timeline for our community work.}
  \label{fig:timeline}
\end{center}
\end{figure*}

Masader Plus was not possible without the help of the community, especially in adding the datasets. In Figure \ref{fig:timeline}, we highlight the timeline. In the next two subsections, we discuss the main events that helped in reaching our current state.

\subsection{Contribution Form}

In the previous version of Masader  \cite{alyafeai2021masader}, two Google Sheets were maintained to track the community input and add it into the catalogue. One of these sheets contains the final reviewed version of the dataset. The other sheet is publicly available for community contribution. The rationale behind this transparency is that anyone who contributes an addition to the catalogue can learn from any prior similar entry.

Despite its simplicity, this data entry strategy is cumbersome to track. First and foremost, the authors must keep track of two sheets. Second, because there is no description for each metadata attribute, it may be unclear to new contributors. There is also a limited flexibility for fields with predefined options (e.g, dialect field). Furthermore, due to the lack of automation, this strategy is prone to errors.

Masader Plus is being developed in collaboration with a developer community. It provides a more accessible way for the community to contribute, as well as for dataset developers to maintain their contributions. In this release,  a Google Contribution Form is enhanced with the following improvements: a) a description for each metadata field; and b) a drop-down list with multiple-options fields.
From the perspective of dataset contributors, they are notified about the status of their submission to Masader. This form was first utilized as part of a two-hour community collaboration hackathon (see Section \ref{community_hackathon}), which resulted in Masader being updated with more than 100 additional datasets. The hackathon demonstrated the effectiveness of our pipeline.

\subsection{Community Hackathon}
\label{community_hackathon}

Using an initial list of datasets that we developed, we invited Arabic NLP students, researchers, and practitioners to join us in a hackathon to add dataset metadata to a Google sheet. We organized the hackathon in collaboration with HuggingFace who supported our initiative. Further, the response from the community to the initiative was very encouraging and we had more than 100 people joining our discord server in one week. After explaining to our community the objectives of the hackathon, we launched the hackathon on the \nth{10} of June 2022 and for two hours the majority of datasets were added. But we allowed anyone to keep adding or modifying  their entries till the \nth{20} of June 2022. At the end of the hackathon, 145 new datasets were added. 

\section{Ethics/Broader Impact Statement}
\label{sec:eth}

Masader began as part of the Big Science initiative\footnote{\url{https://bigscience.huggingface.co/}}. Since its release, it has ignited the interest and support of the Arabic NLP community.  As a result, volunteers, dataset authors, and dataset developers have added more datasets into Masader. Throughout the Masader timeline, we tried to make our process as transparent as possible; by maintaining documentations, engaging with Arabic dataset stakeholders, participating in regular community meetings, and giving the users the option to report instances where metadata is erroneous or incomplete.  By adopting an open and transparent approach, more contributors are joining, leading to growth in both the project's architecture and its content.

Currently, Masader only supports publicly accessible datasets; Due to licensing, the platform does not host the datasets; rather, it simply aggregates the metadata and offers insights.

\section{Discussion and Conclusion}
\label{sec:conc}
In this paper, we presented Masader Plus an interface for exploring over 500 Arabic NLP datasets. We addressed the system description, the design choices and the various features of the interface. We also discussed the significance of community contributions to the Masader Plus experience. This presented version of Masader Plus is considered as version \texttt{1.0.0}, and we intend to make considerable improvements with the following releases. To keep track of updates, we maintain a  changelog\footnote{\url{https://arbml.github.io/masader/changelog}} on our website.

Our primary goal of version \texttt{2.0.0} is to enable seamless integration between the Contribution Form and Google Sheet, as well as to improve the admin and user experience by simplifying the review process and enabling the addition of private datasets respectively.

\section*{Acknowledgements}
We would like to thank Nizar Habash for the helpful conversations. Also, we would like to thank the community for adding new datasets: Abdelrahman Kaseb, Mourad Mars, Abderrahmane Issam,  Afrah Altamimi, Ibrahim Abu Farha, Tarek Eldeeb,  Abdullah Alsaleh, Abdulrahman Kamar, Ahmed Ruby, Aljoharah AlRasheed, Amr Keleg, Bashar Alhafni, Fatima Haouari, Iskander Gaba, Khalid Almubarak, Mohammad Al-Fetyani, Nizar Habash, Nora Alturayeif, Reem Suwaileh, Rua Ismail, Shatha Hakami, Yonatan Belinkov and Saad Benjelloun.

\bibliography{anthology}
\bibliographystyle{acl_natbib}




\end{document}